\documentclass{article}

\usepackage[preprint]{neurips_2026}
\usepackage{siunitx}
\usepackage[utf8]{inputenc} 
\usepackage[T1]{fontenc}    
\usepackage{hyperref}       
\usepackage{url}            
\usepackage{booktabs}       
\usepackage{amsfonts}       
\usepackage{nicefrac}       
\usepackage{microtype}      
\usepackage{xcolor}         
\usepackage{multirow}
\usepackage{graphicx}
\usepackage{amsmath}
\usepackage{caption}
\title{Your Self-Supervised Projection Head Captures Object Co-Occurrence Statistics}

\author{%
    Arthur Aubret \\ Institut Pascal, CNRS, Clermont Auvergne INP \\ Aubière, 63178, France \\
    \texttt{arthur.aubret@uca.fr} 
    \And
    Jochen Triesch \\ Frankfurt Institute for Advanced Studies \\ Ruth-Moufang-Straße 1, Frankfurt am Main, 60438, Germany \\ \texttt{triesch@fias.uni-frankfurt.de} \\
    \And
    Céline Teulière \\ Institut Pascal, CNRS, Clermont Auvergne INP \\ Aubière, 63178, France \\ \texttt{celine.teuliere@uca.fr} \\
}

\begin{document}

\maketitle

\begin{abstract}
 
Self-Supervised Learning (SSL) has achieved impressive success in learning semantic visual representations, yet the underlying principles driving this success remain underexplored. In this work, we hypothesize that common SSL pretext tasks implicitly model object co-occurrence statistics, a fundamental cue for visual learning. To test this, we curate three datasets of segmented objects from existing vision benchmarks using a state-of-the-art segmentation model. Through experiments across many SSL models, we reveal a hierarchical encoding of semantic information: while the visual backbone captures object categories, the projection head specializes in encoding co-occurrences between object categories. As a result, the projection head outperforms the backbone in aligning with human judgments of inter-category similarity. Furthermore, by controlling co-occurrence patterns during pre-training, we demonstrate that encoding object co-occurrences can significantly accelerate the emergence of category-level representations. Our findings uncover a previously hidden learning principle in SSL and suggest a path toward designing more effective pretext tasks by explicitly leveraging object co-occurrence structure.
\end{abstract}
\section{Introduction}

Recent advances in Self-Supervised Learning (SSL) have made it possible to train vision models without annotation, achieving strong performance on downstream tasks such as categorization. Most recent SSL methods remarkably hinge on a shared pipeline (or pretext task) \citep{chen2020simple,caron2021emerging,grill2020bootstrap}: a visual backbone extracts visual features, which feed a multi-layer projection head to learn embeddings invariant to geometric (small random crops, etc.) or color transformations (hue, etc.) of images. There is strong empirical evidence that geometric transformations and the projection head are crucial components of SSL \citep{chen2020simple,caron2020unsupervised}. Although these are known to support position and invariance properties in visual embeddings \citep{tian2020makes,ericsson2022self,chen2020simple}, the underlying principles by which these components support category-wise representation learning remain poorly understood.

A foundational design principle of large language models is to model the statistical structure of word co-occurrences \citep{mikolov2013efficient,radford2019language}. Combined with large-scale datasets, this principle supports the learning of concept representations that capture the semantic structural aspects of our world. In contrast, the established assumption in SSL is that data-augmentations should preserve the task-label information \citep{tian2020contrastive}. For example, two random image croppings should show different parts of the same object. However, vision SSL models may also implicitly encode natural object co-occurrence statistics. Pre-training images often reflect real-world object co-occurrence patterns \citep{turini2022hierarchical}, such as the simultaneous presence of ``food'' and ``forks'' in a dining scene. Random cropping frequently simulates a zoom-in on different objects (\textit{e.g.} ``food'', ``fork'') within a scene. Consequently, SSL models may align embeddings of objects that often co-occur together. This leads to a critical question: Do common SSL methods implicitly capture object co-occurrence statistics, and if so, how does this shape representation learning? To the best of our knowledge, this work is the first to explicitly investigate this hypothesis.

To answer this question, we construct three datasets of segmented objects by leveraging existing large-scale vision datasets and a state-of-the-art segmentation model. This allows us to remove object co-occurrences from the visual inputs, thereby keeping one category of objects per image. Based on this data, we analyze whether internal layers of dozens of SSL methods capture natural object co-occurrence statistics, \textit{e.g.} whether ``food'' and ``fork'' are aligned. To explore the causal impact of such a statistical structure on SSL, we systematically control object co-occurrence statistics during pre-training and quantify their impact on downstream categorization performance.

Our experiments reveal that common SSL models hierarchically encode semantic information: the projection head excels at capturing object co-occurrence statistics, while the visual backbone specializes in discriminating object categories. As a consequence, the projection heads significantly outperform the visual backbones for capturing human inter-category similarity judgments. Finally, we provide experimental evidence that encoding object co-occurrences can drastically improve downstream categorization. Together, these findings uncover a hidden learning principle in SSL: the projection head leverages natural co-occurrence statistics to learn strong visual representations. Our contributions are as follows:

\begin{itemize}
    \item We analyze how object co-occurrences shape self-supervised visual representations learning.
    \item We show that projection heads contain the most effective layers for encoding object co-occurrences.
    \item We discover that encoding object co-occurrences allows the projection head to outperform the backbone in capturing inter-category human similarity judgments.
    \item We provide evidence that encoding object co-occurrences enhances feature extraction for downstream categorization.
\end{itemize}
More broadly, our work suggests that projection heads should not be overlooked after pre-training and opens new avenues for designing pretext tasks that explicitly leverage object co-occurrence structure. The code will be publicly released upon acceptance.

\section{State of the art}

\subsection{Self-Supervised Learning}

Most instance-discrimination SSL methods learn by aligning visual embeddings extracted from two views of an image, where each view results from the original image going through two augmentation pipelines \citep{giakoumoglou2024review}. Augmentations pipelines typically include cropping, flipping or coloring images. The main difference across SSL models lie in how they prevent representational collapse, \textit{i.e.} learning the exact same embedding for all images. Seminal contrastive learning methods explicitly promote dissimilarity between all embeddings \citep{chen2020simple,he2020momentum, wang2022contrastive, hu2021adco,wang2023caco,dwibedi2021little}. Close to contrastive learning, several approaches strive to explicitly decorrelate features within and across different views \citep{zbontar2021barlow,bardes2022vicreg,wang2023self}, with approaches bridging contrastive and decorrelation SSL \citep{zhang2024matrix,liu2022self}. Another set of competitive methods consider asymmetric neural network architectures for processing two views \citep{grill2020bootstrap,chen2021exploring, gidaris2020learning,caron2021emerging,chen2021empirical}. More recently, LeJEPA \citep{balestriero2025lejepa} proposes to enforce an isotropic Gaussian distributed on embedding features, and other methods employ masked modeling to improve upon prior work \citep{assran2023self,zhouimage,oquab2024dinov2,simeoni2025dinov3}. Rather than building a state-of-the-art SSL model, the objective of this work is to understand the principles underpinning the success of these methods.

\subsection{Analysis of the SSL projection head}

The empirical importance \citep{chen2020simple} of the projection head has attracted attention as 1- the projection head is a key architectural difference compared to supervised learning \citep{griggself} and 2- the best performance on downstream tasks is obtained when it is fully discarded \citep{jingunderstanding,chen2020simple,cosentino2022toward,bardes2022vicreg}. One reason for the latter is that the projection head is too specialized for augmentation invariance, a property misaligned with typical downstream tasks \citep{griggself,bordesguillotine,xueinvestigating}. In line with this finding, several works have found that the rank of projection embeddings decreases, compared to that of the backbone representations, which is known as dimensional collapse \citep{gupta2022understanding,jingunderstanding,gui2023unraveling,garrido2023rankme,hua2021feature}. A recent analysis further suggests that the projection head filters out information that is irrelevant for the SSL objective \citep{ouyangprojection}. With spectral analysis, \citet{balestriero2022contrastive} found that SSL optimizes embeddings such that their main singular values match those of the inter-sample similarity matrix. This is consistent with a theory that formulates image augmentations as a graph and posits that the overlap between image augmentations guides SSL \citep{zhang2025augmentation,wangchaos,haochen2021provable}. Visual learning was later reformulated as an interplay between augmentation overlap and architectural inductive biases \citep{cabannes2023ssl,saunshi2022understanding}. In contrast to these works, our work analyzes what is concretely encoded in projection heads and how it shapes learning.

\subsection{Learning Object Co-occurrences}

Visual representations that match object co-occurrence statistics are known to emerge in both humans\citep{turini2022hierarchical,bonner2021object} and supervised learning models\citep{bracci2023representational}. In SSL, some studies have assumed the importance of spatial co-occurrences between object features \citep{aubret2026seeing,caron2020unsupervised}. However, few studies have investigated the presence of representations capturing co-occurrences between distinct objects in SSL. \citet{aubret2024learning} found a hierarchical encoding of semantic information in SSL, but they manually defined object co-occurrences. In contrast, we focus our analysis on natural co-occurrences present in the standard training regime. A more recent study \citep{schaumloffel2026temporal} found that learning temporally consistent representations from human egocentric videos promotes the encoding of objects' context. Yet, they neglect the projection head. To the best of our knowledge, we are the first to show that encoding natural object co-occurrences is a key principle of common visual SSL approaches.

\section{Method}

\subsection{Motivation}\label{sec:motivation}

Our working hypothesis presumes that the SSL crop-based augmentation pipeline frequently yields pairs of images depicting different objects. We start by testing this assumption on ImageNet-1k (I1K), the most established dataset for visual SSL. Since I1K lacks semantic segmentation annotations, we use a pretrained ResNet50 as a proxy to estimate the frequency of semantically inconsistent pairs. We sample 1M of crop-augmented image pairs from the train split and compute 1- an upper bound on inconsistency by measuring how often the predicted class labels for a pair differ and 2- A lower bound by counting how frequently each crop's predicted label falls outside the other's top-5 predictions.

\begin{figure}
    \centering
    \includegraphics[width=1\linewidth]{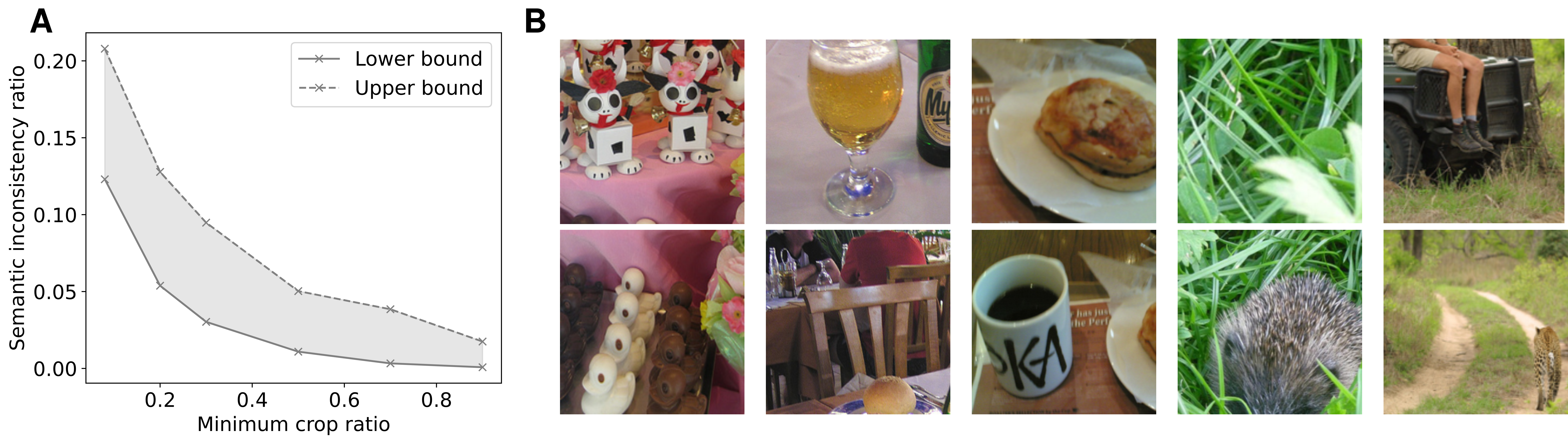}
    \caption{A) Ratio of semantically inconsistent pair of crops when varying the minimum crop ratio of SSL's augmentation pipeline. B) Examples of semantically inconsistent pairs obtained with random cropping.}
    \label{fig:motivation}
\end{figure}

As shown in Figure~\ref{fig:motivation}A, augmented views are semantically inconsistent approximately 15\% of the time under standard hyperparameters (\textit{e.g.}, a minimum crop size of 0.08). Furthermore, Figure~\ref{fig:motivation}B reveals that cropping can align co-occurring objects from diverse categories, such as toys, dining objects (chair and glass, mug and food) or meadow-related elements (grass and hedgehog). These observations motivate our investigation into whether SSL models implicitly learn from such co-occurrence patterns.

\subsection{Datasets and metrics}\label{sec:datasets}

\paragraph{Datasets and object co-occurrence matrices.}
We want to assess whether and how SSL methods encode object co-occurrence statistics. We start by deriving a dataset for which each image contains only a single object, devoid of contextual information. This setup ensures that any contextual relationships between objects must emerge from statistical patterns discovered during training. To do so, we use the THINGS dataset \citep{hebart2023things}, which contains 26,107 naturalistic images distributed in 1,854 object concepts, with each image depicting a single, clearly visible object. To further remove residual background information, we preprocess all images with Lang-SAM \citep{langsam}, a language-conditioned segmentation pipeline. Lang-SAM combines Grounding-DINO \citep{liu2023grounding} to generate concept-specific bounding boxes and SAM-2 \citep{ravi2024sam2} to segment the objects within these boxes. We manually filter out poorly segmented images. Additionally, we crop and extract segmented objects larger than $32 \times 32$ pixels from the LVIS validation split \citep{gupta2019lvis} as an alternative dataset. 

To test whether SSL models encode contextual information in segmented objects, we first estimate natural co-occurrence statistics. We use three large-scale scene datasets that provide semantic annotations of several objects within each image: the COCO training and validation splits, the ADE-20K training split, and the LVIS training split. For each dataset, we compute a co-occurrence matrix by counting the total number of times each pair of object concepts appears together. To align these statistics with our segmented object datasets, we retain only overlapping classes between each scene dataset and its corresponding segmented object dataset. This yields three paired datasets (Co-occurrence dataset-Segmented dataset):
\begin{itemize}
    \item COCO-THINGS: 736 segmented THINGS images across 53 categories,
    \item ADE20K-THINGS: 918 segmented THINGS images across 77 categories,
    \item LVIS-LVIS: 118,068 extracted LVIS object images across 1,002 categories.
\end{itemize}

\paragraph{Representational Similarity Analysis.}

We perform a Representational Similarity Analysis (RSA) to quantitatively assess whether layer representations reflect natural co-occurrence matrices. For each layer of a model, we average the representations across segmented images that belong to the same category and compute the pair-wise cosine representational similarity matrix (RSM) between category representations. Then, we compute the RSA Kendall's tau between the upper triangle of the RSM and the co-occurrence matrix.

\paragraph{Context classification accuracy}

As an alternative to RSA, we propose a context classification accuracy metric to measure the alignment between object co-occurrences and visual representations. For COCO-THINGS, we manually assign its 53 object categories to 10 spatial contexts in \{``Urban street'', ``Countryside'', ''Savanna'', ``Sea'', ``Snow mountain'', ``Sport field'', ``Kitchen'', ``Bathroom'', ``Living room'', ``Bedroom''\}. For ADE20k-THINGS and LVIS-LVIS, we run three times a spectral clustering algorithm \citep{von2007tutorial} on the normalized logarithm of the co-occurrence matrix. Each clustering provides 10 clusters, which we label as co-occurrence contexts. In \sectionautorefname~\ref{sec:hierarchy}, we experiment varying the number of clusters.

To compute an accuracy, we first build category representations by averaging the representations across segmented images that belong to the same category, as in \sectionautorefname~\ref{sec:datasets}. Then, we generate a series of tuples composed of one randomly sampled category representation, a random category representation belonging to the same context, and a random category representation from each other context. Then, we compute how often the first representation is closer to that of the same context than to representations from other contexts. For each context assignment, we assess 1,000/10,000 tuples for *-THINGS/LVIS-LVIS and report the average accuracy over tuples and context assignments. In \appendixautorefname~\ref{app:imagelevel}, we also consider image-level comparison metrics.

\subsection{Pretrained models}\label{sec:pretrainedmodels}

To emphasize the general representational principles of SSL, we evaluate a diverse set of SSL models pretrained on Imagenet-1k, including their projection heads. Our selection spans multiple SSL paradigms (contrastive, decorrelation, distillation) and neural architectures (ViT-B/16, ResNet50), specifically: MoCoV3 with ViT-B/16 and Resnet50, BYOL with Resnet50, VICReg with Resnet50 and DINO with ViT-B/16. In some experiments, we also pretrain SSL models, and use other pretrained models, as detailed in \appendixautorefname~\ref{app:newlypretrained}.

\section{Results}

\subsection{The Projection Head Captures Natural Object Co-occurrences}\label{sec:cooccur}

Here, we investigate whether the projection head of classical SSL methods learns representations that capture object co-occurrence statistics. In \tableautorefname~\ref{table:cooc}, we observe that an internal layer of the projection head consistently outperforms the final layers of the backbone in encoding these statistics across all datasets and SSL methods. Minor exceptions are BYOL-R50 on LVIS-LVIS (RSA Kendall's tau) and MoCoV3-ViT-B on COCO-THINGS (context classification), which remain unexplained. Early and intermediate layers of the projection head are often particularly suited to capture object co-occurrence structure, while the final linear layer tends to perform poorly. These results are confirmed with a broader set of pretrained models in \appendixautorefname~\ref{app:controlled} and with more backbone layers in \appendixautorefname~\ref{app:morehidden}. When comparing across methods, DINO-ViT-B stands out as the best model for capturing object co-occurrence statistics, with a single exception for Kendall's tau in LVIS-LVIS. As DINO is also the best performing model for ImageNet classification, this suggests that capturing object co-occurrences may correlate with better image representations for downstream classification. We further explore this hypothesis in \sectionautorefname~\ref{sec:boostcategory}.

\begin{figure}
    \centering
    \includegraphics[width=1\linewidth]{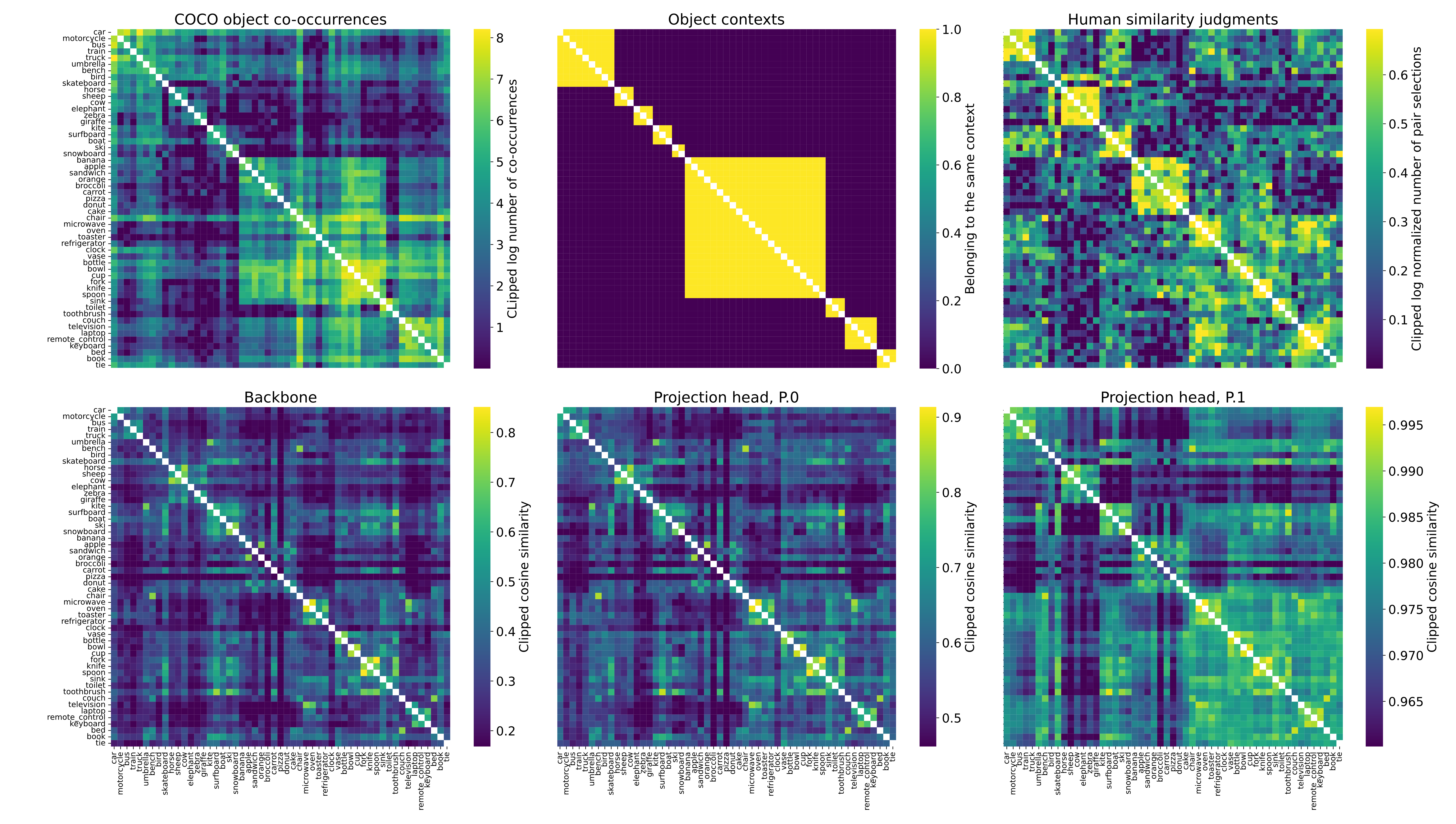}
    \caption{Top row: RSMs for COCO object co-occurrences, context assignments, and human similarity judgments. Bottom row: RSMs for visual representations learnt by DINO-ViT-B after the visual backbone, the first and the second linear layer of the projection head (P.0 and P.1, respectively). To avoid outliers to bias the visualization, we lower bound each matrix by the first decile of the matrix.}
    \label{fig:rsms}
\end{figure}

To better understand the semantic aspects captured by the projection head, we visualize in \figureautorefname~\ref{fig:rsms} the RSMs of object co-occurrences, representation layers and clusters of contexts. First, we observe that the hand-defined contexts roughly capture the geometry of object co-occurrences. Differences are noticeable, which potentially stem from the difference between COCO and naturalistic environments or imperfect context construction. The backbone of DINO-ViT-B captures only small clusters of object co-occurrences, such as sea-related (kite, surfboard, boat) or dining-related (fork, knife, spoon) objects. In contrast, projection layer P.1 further reflects broader semantic contexts such as in-door household objects or food-related items. The intermediate projection layer P.0 shows in-between behavior. 

These results demonstrate that layers of projection heads learn a representational structure aligned with object co-occurrence statistics. However, is the projection head necessary for representing object co-occurrences? To answer this question, we pretrain SSL model while modifying the projection head (cf. \appendixautorefname~\ref{app:newlypretrained}). In \tableautorefname~\ref{tab:withwithout}, we compare co-occurrence encoding metrics with and without a non-linear projection head during pre-training. We observe that the projection head consistently improves co-occurrence encoding across datasets, methods and metrics. There is a single exception for the ADE20k-THINGS context accuracy of SimCLR. Taken together, these findings demonstrate that the projection head is crucial for encoding object co-occurrences. 

\begin{table}
\centering
\caption{Evaluation of the backbone (B) and internal layers of projection heads (P.n) based on whether they capture the co-occurrence statistics of objects. ``mP.n'' refers to the maximal value within the n-th block ``Linear, BatchNorm, ReLU'' of the projection head. ``P.n'' refers to the n-th linear layer of the projection head. ``na'' occurs when the pretrained model does not contain the n-th linear layer. C-T, A-T and L-L refer to COCO-THINGS, ADE20k-THINGS and LVIS-LVIS, respectively. 
}
\label{table:cooc}
\begin{tabular}{c|c|cccc|cccc}
\toprule
\multirow{2}{*}{Cooc.} & \multirow{2}{*}{Method} & \multicolumn{4}{|c|}{RSA Kendall's tau} & \multicolumn{4}{|c}{Context accuracy} \\
\cmidrule(lr){3-6} \cmidrule(lr){7-10}
&  & B & mP.0 & mP.1 & mP.2 & B & P.0 & P.1 & P.2 \\ 
\midrule
&MoCoV3-R50 & 1.5 & \textbf{13.7} & 6.4 & na & 39.2 & \textbf{47.7} & 25.5 & na  \\
&BYOL-R50 & 3.5 & \textbf{22.3} & -4.9 & na  & 38.6 & \textbf{42.0} & 23.4 & na  \\
C-T&VICReg-R50 & 2.6 & 14.7 & \textbf{16.8} & -3.4 & 35.1 & \textbf{40.8} & 39.3 & 22.4 \\
 &MoCoV3-ViT-B & 6.5 & 7.1 & \textbf{14.8} & 5.9 & \textbf{39.9} & 38.9 & 36.2 & 33.0  \\
&DINO-ViT-B & 9.0 & 21.5 & \underline{\textbf{33.4}} & 13.9 & 43.5 & 45.7 & \underline{\textbf{50.3}} & 40.7 \\
\midrule
&MoCoV3-R50 & 14.1 & 16.9 & \textbf{17.7} & na  & 22.0 & \textbf{23.6} & 22.6 & na \\
&BYOL-R50 & 9.4 & \textbf{11.3} & 8.0 & na  & 23.2 & \textbf{26.8} & 19.5 & na \\
A-T&VICReg-RN0 & 11.2 & \textbf{17.2} & 16.3 & 11.6  & 24.2 & \textbf{26.6} & 24.6 & 25.7 \\
&MoCoV3-ViT-B & 15.8 & \textbf{18.6} & 17.0 & 14.8 & 27.0 & 28.1 & \textbf{28.2} & 24.9  \\
&DINO-ViT-B & 19.3 & 21.0 & 20.5 & \underline{\textbf{22.4}} & 22.8 & 23.6 & 26.7 & \underline{\textbf{33.0}} \\
\midrule
&MoCoV3-R50 & 40.7 & \textbf{45.7} & 13.2 & na  & 19.0 & \textbf{20.2} & 14.7 & na  \\
&BYOL-R50 & \textbf{39.8} & 39.3 & 17.9 & na  & 18.5 & \textbf{20.6} & 13.3 & na  \\
L-L&VICReg-R50 & 37.1 & 40.3 & \underline{\textbf{49.0}} & 16.2  & 19.0 & \textbf{19.7} & 18.6 & 16.1   \\
&MoCoV3-ViT-B & 43.8 & \textbf{44.0 }& 43.0 & 20.2 & 18.9 & \textbf{19.6} & 19.5 & 18.4  \\
&DINO-ViT-B & 39.8 & \textbf{43.3} & 40.6 & 27.5  &21.4 & 21.1 & \underline{\textbf{23.9}}& 22.6 \\
\bottomrule
\end{tabular}
\end{table}

\begin{table}[]
    \centering
        \caption{Comparison of SSL models with and without non linear projection heads, for encoding object co-occurrences. In the ``without'' condition, the projection head is a linear layer (two for DINO with the final expansion layer). In the ``with'' condition, we use the setting from the original paper (One hidden layer for SimCLR and BYOL, two for DINO). C-T, A-T and L-L refer to COCO-THINGS, ADE20k-THINGS and LVIS-LVIS, respectively.}
    \label{tab:withwithout}
    \begin{tabular}{c|c|SS|SS}
        \toprule
        \multirow{2}{*}{Cooc.} & \multirow{2}{*}{Method} & \multicolumn{2}{|c|}{RSA Kendall's tau} & \multicolumn{2}{|c}{Context accuracy} \\
\cmidrule(lr){3-4} \cmidrule(lr){5-6}
&  & without & with & without & with \\ 
\midrule

         & SimCLR & 4.3 & \textbf{14.2} & 36.6 & \textbf{42.4}   \\
        C-T & BYOL & 2.3 & \textbf{22.5} & 37.6 & \textbf{46.8}  \\
         & DINO & 12.8 & \textbf{21.4} & 36.9 & \textbf{41.7}   \\
        \midrule
         & SimCLR & 12.4 & \textbf{19.7} & \textbf{24.5} & 22.6 \\
        A-T & BYOL & 13.8 & \textbf{16.4} & 22.3 & \textbf{23.1} \\
         & DINO & 16.1 & \textbf{19.9} & 26.4 & \textbf{26.7} \\
         \midrule
         & SimCLR & 26.9 & \textbf{38.3} & 18.3 & \textbf{20.9} \\
        L-L & BYOL & 26.3 & \textbf{40.7} & 17.5 & \textbf{20.6} \\
         & DINO & 39.3 & \textbf{42.9} & 18.8 & \textbf{20.7} \\
        \bottomrule
        
    \end{tabular}

\end{table}

\subsection{The Projection Head Better Captures Human Object Similarity Judgments}\label{sec:judgements}

If the projection head captures semantic information beyond object categorization, it may reflect human inter-category similarity judgments. To test this hypothesis, we leverage the original THINGS dataset, which comprises $\approx 4\text{M}$ odd-one-out human judgments over image triplets (train set). We construct a Representational Similarity Matrix (RSM) where each entry reflects the frequency with which a pair of categories was selected by participants as the most similar within the triplet, normalized by the number of times that pair appeared in a triplet. Then, we perform an RSA by measuring the correlation between the human RSM and other RSMs (e.g., encode representations or co-occurrence statistics). First, to assess the plausibility of our hypothesis, we compare human semantic similarity and object co-occurrence statistics. We find a positive Kendall's Tau of 0.266 and a Spearman correlation score of 0.374, confirming that object co-occurrences correlate with human similarity judgments. Building on this, we compare the organizational structure of visual representations in \tableautorefname~\ref{tab:humanrsa} with that of human RSM. We consistently see that at least one layer of the projection head better encodes inter-category human similarity judgments than the final layer of the backbone.

Next, we investigate whether the projection head's superior performance in human similarity judgments stems from its ability to encode object co-occurrences. We gather 14 publicly pretrained SSL models as well as 12 SSL models pretrained on ImageNet-1k for 100 epochs with different numbers of projection layers (see \appendixautorefname~\ref{app:newlypretrained} for details). This yields a total of 101 representation layers, covering the final layer of the backbones and the internal layers of the projection heads. As seen in  \figureautorefname~\ref{fig:correlation}, representation layers that excel at encoding object co-occurrences tend to perform well in capturing human similarity judgments. A strong Pearson correlation supports this observation with THINGS-COCO ($r=0.52$, $p=\num{1.72e-8}$), ADE20k-THINGS ($r=0.47$, $p=\num{6.14e-7}$) and LVIS-LVIS ($r=0.39$, $p=\num{6.19e-5}$). Taken together, these results suggest that the projection head's ability to encode object co-occurrences makes it more suitable than the visual backbone for modeling inter-category human similarity ratings.

    \begin{minipage}{0.53\textwidth}
    \centering
        \captionof{table}{RSA Kendall's tau between human inter-category similarity judgments and the visual representations of layers in a SSL model. ``P.n'' refers to the maximal value within the n-th block ``Linear, BatchNorm, ReLU'' of the projection head.}
        \label{tab:humanrsa}
        \begin{tabular}{l|cccc}
        \toprule
        Method & B & P.0 & P.1 & P.2 \\
        \midrule
        MoCoV3-50 & 17.8 & \textbf{19.8} & 9.0 & na  \\
        BYOL-R50 & 18.3 & \textbf{22.7} & 8.0 & na \\
        VICReg-R50 & 17.7 & 21.0& 13.4 & \textbf{23.5} \\
        MoCoV3-ViT-B & 18.1 & 19.0 & \textbf{22.9} & 21.6 \\
        DINO-ViT-B & 30.5 & 28.5 & \textbf{31.1} & 28.0 \\
        \bottomrule
        \end{tabular}

    \end{minipage}
        \hfill 
    \begin{minipage}{0.42\textwidth}
        \centering
        \includegraphics[width=1\textwidth]{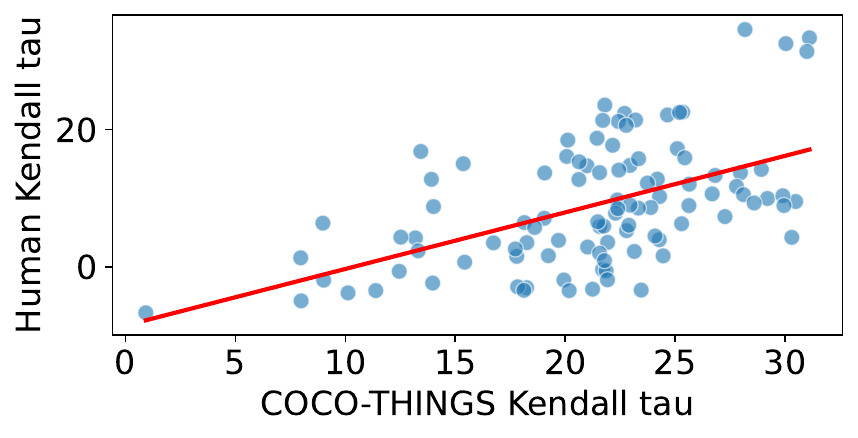}
        \captionof{figure}{Correlation between encoding COCO-THINGS object co-occurrences and THINGS human similarity judgments, with all pretrained models described in \sectionautorefname~\ref{sec:pretrainedmodels}.}
        \label{fig:correlation}
    \end{minipage}
    
\subsection{Extracting Object Co-occurrences Boosts Category Learning}\label{sec:boostcategory}

\begin{figure}
    \centering
    \includegraphics[width=1\linewidth]{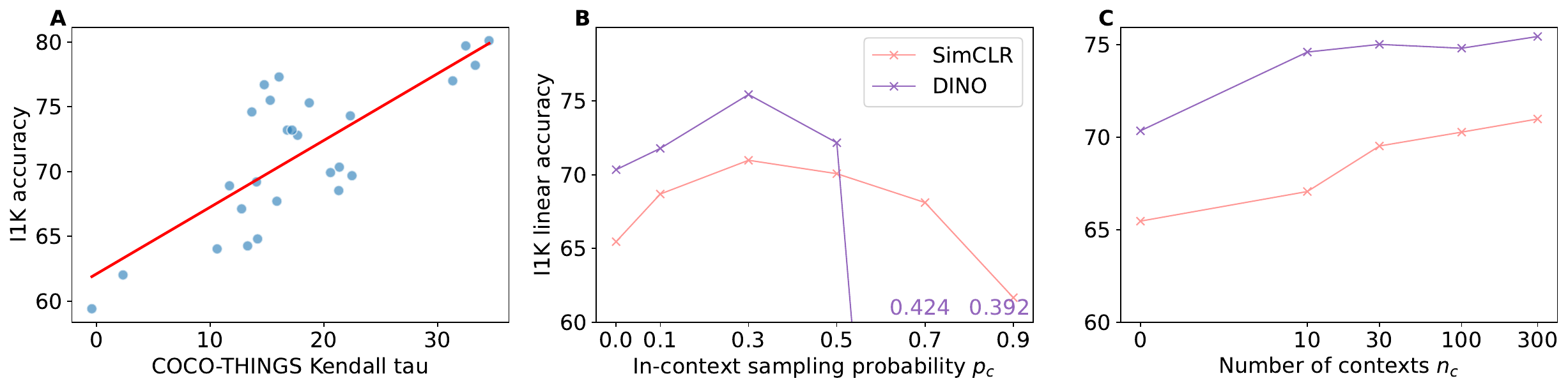}
    \caption{A) Correlation between Imagenet-1k accuracy and the maximum of COCO-THINGS Kendall's tau in the projection head. We plot the Least-Square regression in red. We consider the full set of pretrained models described in \sectionautorefname~\ref{sec:pretrainedmodels}.
    B) Imagenet-1k accuracy while varying the probability $p_c$ of sampling two different images from the same context. We use $n_c=300$. Due to DINO's collapse for $p=0.7$ and $p=0.9$, we display the raw accuracies for these experiments. C) Imagenet-1k accuracy against the number of contexts. We use $p_c=0.3$.}
    \label{fig:booster}
\end{figure}

Here, we investigate whether learning to encode object co-occurrences enhances category learning in SSL. First, we test whether encoding object co-occurrences in the projection head correlates with improved visual features for downstream categorization. In \figureautorefname~\ref{fig:booster}A), we clearly see an increase in category recognition with higher COCO-THINGS Kendall's tau. This is quantitatively confirmed by Pearson scores, with co-occurrences from COCO-THINGS ($r=0.78$, $p=\num{7.95e-6}$), ADE20k-THINGS ($r=0.48$, $p=\num{1.81e-2}$) and LVIS-LVIS ($r=0.79$, $p=\num{4.84e-6}$).

Correlation does not necessarily imply causation. To establish a causal link, we define new controlled co-occurrence statistics during training. We uniformly and randomly distribute I1K classes into $n_c$ co-occurrence contexts. The higher $n_c$, the lower the number of classes and images per context. For example, boats and elephants might belong to the same context, while forks and spoons might not. During training, we sample two images for each positive pair. Unlike usual SSL, we sample with probability $p_c$ the second image from a different class within the same context, as the first image. Then, systematically varying $p_c$ and $n_c$ allows us to explore how co-occurrence statistics alone shape category learning in SSL. We pretrain two methods (SimCLR, DINO) for 100 epochs under these controlled conditions. In \figureautorefname~\ref{fig:booster}B), we see a sweet spot at $p_c=0.3$, with a drastic accuracy improvement of up to +5.52\% compared to the baseline with no hand-crafted contexts ($p_c=0$). This indicates that co-occurrence statistics can play a substantial role in shaping visual learning. However, the consistent accuracy drop for $p_c > 0.3$ suggests that alignment of same-image augmentations remains a critical learning signal beyond co-occurrence encoding. In \figureautorefname~\ref{fig:booster}C), we further find that increasing the number of contexts $n_c$ yields enhanced visual representations, compared to the baseline without hand-crafted contexts ($n_c=0$). This means that small clusters of repeated co-occurrences, such as one composed of ``sink'', ``toilet'' and ``toothbrush'' (cf. \figureautorefname~\ref{fig:rsms}),  are particularly effective for co-occurrence-based SSL. Taken together, these results show that repeated object co-occurrences enhance self-supervised representation learning.

\subsection{SSL Hierarchically Extracts Semantic Representations}\label{sec:hierarchy}

We found in \sectionautorefname~\ref{sec:cooccur} that the projection head extracts representations that capture object co-occurrence statistics. One possible explanation is that the projection head simply learns better visual representations for capturing diverse semantic information.
To verify this, we define the in-context one-shot category classification as follows: we randomly sample an image as well as an image from each category in our context assignments. Then, we assess how frequently an image representation is closer to that of the same category than other categories. We show results only for THINGS-COCO and LVIS-LVIS in \tableautorefname~\ref{tab:incontext}, but consistent results are obtained with THINGS-ADE20k (cf. \appendixautorefname~\ref{app:hierarchy2}). We observe that the performance of the projection consistently decreases with layer depth across all datasets and methods. This demonstrates that SSL models hierarchically organize visual representations, transitioning from categorical to contextual organization in deeper layers.

This result raises the question of whether semantic information is also hierarchically organized within the projection head. Object co-occurrences are naturally hierarchical \citep{turini2022hierarchical}. For example, spoons may co-occur very frequently with plates, frequently with bananas, regularly with teapots, occasionally with a television etc. Different layers of the projection head may capture these varying levels of semantic co-occurrences. To investigate this, we define five levels of contextual organization by varying the number of clusters of LVIS categories between \{3,10,30,100,300\}. Then, we re-train various SSL models for 100 epochs while controlling the number of projection layers (cf. \appendixautorefname~\ref{app:newlypretrained}). This allows us to compute the context classification accuracy for each layer and contextual level. In \tableautorefname~\ref{tab:incontext}, the results confirm that the projection head consistently outperforms the backbone for encoding object co-occurrences. However, we do not see consistent trends regarding the best layer across levels of contextual organization. This indicates that additional mechanisms might be necessary to enforce a more explicit hierarchical encoding of object co-occurrence statistics within the projection head.

\begin{table}
\centering
\caption{In-context one-shot category classification accuracy.}
\label{tab:incontext}
\begin{tabular}{c|ccccc|ccccc}
\toprule
\multirow{2}{*}{Method} &  \multicolumn{5}{|c|}{COCO-THINGS} & \multicolumn{5}{|c}{LVIS-LVIS} \\
\cmidrule(lr){2-6} \cmidrule(lr){7-11}
     & B & P.0 & P.1 & P.2 & P.3 & B & P.0 & P.1 & P.2 & P.3 \\
\midrule
MoCoV3-R50 & \textbf{90.0} & 88.7 & 78.6 & na & na & \textbf{81.3} & 74.5 & 64.2 & na & na \\
BYOL-R50 & \textbf{89.9} & 83.1 & 71.7 & na & na & \textbf{80.9} & 76.1 & 65.5 & na & na \\
VICReg-R50 & \textbf{87.4} & 83.9 & 81.1 & 72.8 & na & \textbf{78.1} & 75.6 & 74.4 & 64.5 & na \\
MoCoV3-ViT-B & \textbf{89.7} & 89.6 & 87.1 & 85.0 & na & \textbf{80.2} & 79.9 & 76.0 & 72.5 & na \\
DINO-ViT-B & \textbf{92.9} & 92.3 & 90.3 & 89.3 & 86.0 & \textbf{85.4} & 84.1 & 80.7 & 79.6 & 74.2 \\
\bottomrule
\end{tabular}
\end{table}

\begin{table}
\centering
\caption{LVIS-LVIS Context classification accuracy for several methods and context definitions. }
\label{tab:lvis}
\begin{tabular}{c|c|cccc|ccccc}
\toprule
\multirow{2}{*}{Method} & \multirow{2}{*}{ \# contexts} & \multicolumn{4}{|c}{3 projection layers} & \multicolumn{5}{|c}{4 projection layers} \\
\cmidrule(lr){3-6} \cmidrule(lr){7-11}
& & B & P.0 & P.1 & P.2 & B&  P.0 & P.1 & P.2 & P.3 \\

\midrule
\multirow{5}{*}{SimCLR-R50} & 3 & 40.4 & 42.9 & \textbf{45.4} & 41.5 & 39.9 & 42.2 & 43.2 & \textbf{44.6} & 41.7 \\
 & 10 & 18.1 & 19.8 & \textbf{20.3} & 18.0 & 17.3 & 18.8 & 19.4 & \textbf{19.7} & 17.9 \\
 & 30& 12.5 & 14.3 &\textbf{15.0} & 13.0  & 13.2 & 13.5 & \textbf{14.2} & 13.6 & 12.4  \\
& 100 & 7.7 & 8.7 & \textbf{9.4} & 9.2 & 8.1 & 8.5 & 8.1 & 9.1 & \textbf{10.0} \\
 & 300 & 5.7 & 7.8 & \textbf{9.1} & 8.8  & 5.7 & 5.4 & 6.0 & 6.1 & \textbf{6.7} \\
\midrule
\multirow{5}{*}{BYOL-R50} & 3& 41.7 & 45.1 & \textbf{45.2} & 39.2  & 42.2 & 45.0 & \textbf{44.7} & 43.2 & 41.0  \\
 & 10 & 17.9 & 20.5 & \textbf{20.7} & 16.5 & 17.9 & 20.5 & \textbf{20.7} & 16.5 \\
 & 30 & 14.2 & 16.3 & \textbf{17.4} & 14.9 & 15.3 & 16.5 & 16.2 & \textbf{17.5}& 16.4 \\
 & 100 & 8.5 & \textbf{9.3} & 8.2 & 8.2  & 8.2 & \textbf{9.4} & 8.9 & 8.1 & 8.0 \\
& 300 & 6.5 & \textbf{9.3} & 7.3 & 7.8  & 8.0 & 9.4 & 9.4 & 9.6 & \textbf{10.8} \\
\midrule
\multirow{5}{*}{DINO-ViT-B} & 3 & 36.8 & 43.6 & \textbf{44.0} & 42.3   & 36.6 & 44.4 & 43.4 & \textbf{44.6} & 43.7  \\
 & 10 & 18.6 & 19.5 & \textbf{20.2} & 18.4   & 19.0 & 19.6 & 19.5 & \textbf{20.7} & 18.4 \\
 & 30 & 14.6 & 17.3 & \textbf{17.7} & 15.5  & 13.6 & 15.1 & 16.4 & 17.7 & \textbf{18.4}   \\
 & 100 & 7.6 & 8.8 & 8.7 & \textbf{9.2} & 7.8 & 8.5 & 8.5 & 9.5 & \textbf{10.3} \\
 & 300& 5.7 & \textbf{10.3} & 7.1 & 5.1 & 5.8 & \textbf{9.6} & 8.6 & 9.3 & 6.1  \\
\bottomrule
\end{tabular}
\end{table}

\section{Conclusion}

The learning principles driving visual SSL are poorly understood, the primary assumption being that augmentations should preserve task-label information. Here, we explored the hypothesis that common SSL models also learn by implicitly encoding object co-occurrence statistics. We found that SSL methods trained on Imagenet-1k indeed capture object co-occurrence statistics in their representations. Our analysis further demonstrates that the projection head outperforms the visual backbone in capturing both object co-occurrences and human inter-category similarity judgments. Furthermore, controlled experiments suggest that encoding object co-occurrences supports the learning of category-wise visual representations in the backbone. Overall, our study reveals the joint role of the projection head and object co-occurrences in visual SSL, paving the way for pre-training strategies that explicitly leverage object co-occurrence statistics.

Our analysis opens several avenues for further investigation.  First, we noticed that multiple factors impact object co-occurrence encoding, such as the neural architecture or the SSL method. Understanding how and why these factors shape co-occurrence encoding might yield novel insights for developing SSL models. Second, modern foundational models such as DinoV2 \citep{oquab2024dinov2} do not release the weights of their projection heads. Future works should confirm our findings on these models. Nevertheless, our study unveils object co-occurrence encoding as a hidden, yet important, learning principle in SSL.

\section*{Acknowledgements}
The authors acknowledge the ANR – FRANCE (French National Research Agency) for its financial support of the MeSMERise project ANR-23-CE23-0021-01. This work was funded by the Deutsche Forschungsgemeinschaft: DFG project 5368 (``Abstract REpresentations in Neural Architectures (ARENA)'') and supported by the Deutsche Forschungsgemeinschaft (German Research Foundation, DFG) under Germany’s Excellence Strategy (EXC 3066/1 ``The Adaptive Mind'', Project No. 533717223). The authors gratefully acknowledge the computing time provided to them at the NHR Center NHR@SW at Goethe University Frankfurt (project autolearn). This is funded by the Federal Ministry of Education and Research, and the state governments participating on the basis of the resolutions of the GWK for national high performance computing at universities (www.nhr-verein.de/unsere-partner). This work was also supported in part by computational resources provided by the MaSC high-performance computing cluster at Philipps-Universität Marburg, the Hessian Center For Artificial Intelligence and GENCI-IDRIS (Grant 2026-AD011017150).

\bibliographystyle{unsrtnat}
\bibliography{main_bib}

\appendix
\newpage




\section{Pretrained models}\label{app:newlypretrained}

\tableautorefname~\ref{tab:fullpretrained} shows the full set of publicly available pre-trained models used in our experiments. In \appendixautorefname~\ref{app:controlled} and \sectionautorefname~\ref{sec:hierarchy}, we pretrain for 100 epochs SSL models with a controlled number of projection heads. We use hyper-parameters provided in the main papers and the solo-learn codebase \citep{JMLR:v23:21-1155}, with few modifications. To adapt to the smaller number of epochs, we increase the base learning rate to 1.6 for Resnet50. We always apply the same augmentation pipeline as in BYOL \citep{grill2020bootstrap}; for DINO, we consider two large crops ([0.4, 1]) and four small crops ([0.05, 0.4]). We always report the final online linear accuracies, following \citet{bordes2023towards}. The code and configuration files will be publicly released upon acceptance.

Computational resources required for pre-training the models are described in the original papers. In our experiments, we used 4 NVIDIA A100 for BYOL and 8 AMD Instinct MI210 for SimCLR/DINO. Pretraining SimCLR and BYOL lasted 20 to 30 hours for BYOL/SimCLR, and approximately 72 hours for DINO. Computing the metrics is done in less than 1 hour on four NVIDIA A100.

\begin{table}[]
    \centering
    \caption{Full set of publicly available pre-trained models studied in some of our experiments. The top block shows the models used in all our experiments, while others were used for correlation analyses. *We infer that DINO-R50 was pre-trained for 800 or 1000 epochs from the paper.}
    \label{tab:fullpretrained}
    \begin{tabular}{c|c|c|c}
        \toprule
        Method & Neural network & Pre-training epochs & I1K accuracy \\
        \midrule
            MoCoV3 & Resnet50 & 1000 &  74.6 \\
            BYOL & Resnet50 & 800 & 74.4 \\
            VICReg & Resnet50 & 1000 & 73.2 \\
            MoCoV3 & ViT-B/16 & 300 & 76.7 \\
            DINO & ViT-B/16 & 300 & 78.2 \\
            \midrule
            MoCoV3 & Resnet50 & 300 &  72.8 \\
            MoCoV3 & Resnet50 & 100 &  68.9 \\
            MoCoV3 & ViT-S/16 & 300 & 73.2 \\
            VICReg & Resnet50x2 & 1000 & 75.5 \\
            VICReg & Resnet200x2 & 1000 & 77.3 \\
            DINO & Resnet50 & 800-1000* & 75.3 \\
            DINO & ViT-S/8 & 300 & 79.7 \\
            DINO & ViT-S/16 & 300 & 77 \\
            DINO & ViT-B/8 & 300 & 80.1 \\
         \bottomrule
    \end{tabular}
\end{table}

\section{Extended results}

\subsection{Standard deviation over context accuracy}\label{app:std}

Because of space constraints, we could not include the standard deviation of the context accuracy across context definitions in \tableautorefname~\ref{table:cooc}. \tableautorefname~\ref{table:cooc2} reports the missing standard deviation. Overall, the accuracy varies only slightly over different context assignments.

\begin{table}[]
\centering
\caption{Replicate of the context accuracy reported in \tableautorefname~\ref{table:cooc}, with standard deviation across the three different context assignments.}
\label{table:cooc2}
\begin{tabular}{c|c|cccc}
\toprule
\multirow{2}{*}{Cooc.} & \multirow{2}{*}{Method} & \multicolumn{4}{|c}{Context accuracy} \\
\cmidrule(lr){3-6} 
&  & B & P.0 & P.1 & P.2 \\ 
\midrule
& MoCoV3-R50 & 22.0 $\pm$ 0.94 & \textbf{23.6 $\pm$ 1.37 }& 22.6 $\pm$ 1.65 & na  \\
& BYOL-R50 & 23.2 $\pm$ 1.30 & \textbf{26.8 $\pm$ 1.27} & 19.5 $\pm$ 0.61 & na  \\
A-T & VICReg-R50 & 24.2 $\pm$ 0.88 & \textbf{26.6 $\pm$ 1.08} & 24.6 $\pm$ 0.73 & 25.7 $\pm$ 1.85 \\
& MoCoV3-ViT-B & 27.0 $\pm$ 0.86 & 28.1 $\pm$ 0.95 & \textbf{28.2 $\pm$ 0.84} & 24.9 $\pm$ 1.41  \\
& DINO-ViT-B & 22.8 $\pm$ 1.27 & 23.6 $\pm$ 1.21 & 26.7 $\pm$ 1.54 & \textbf{33.0 $\pm$ 0.22}  \\
\midrule
& MoCoV3-R50 & 19.0 $\pm$ 0.66 & \textbf{20.2 $\pm$ 0.90} & 14.7 $\pm$ 0.41 & n/a  \\
& BYOL-R50 & 18.5 $\pm$ 0.50 & \textbf{20.6 $\pm$ 0.53} & 13.3 $\pm$ 0.59 & n/a \\
L-L & VICReg-R50 & 19.0 $\pm$ 0.51 & \textbf{19.7 $\pm$ 0.51} & 18.6 $\pm$ 0.54 & 16.1 $\pm$ 2.26 \\
& MoCoV3-ViT-B & 18.9 $\pm$ 0.45 & \textbf{19.6 $\pm$ 0.92} & 19.5 $\pm$ 1.14 & 18.4 $\pm$ 0.77  \\
& DINO-ViT-B & 21.4 $\pm$ 1.07 & 21.1 $\pm$ 1.08 &\textbf{ 23.9 $\pm$ 1.14} & 22.6 $\pm$ 1.01 \\
\bottomrule
\end{tabular}
\end{table}

\subsection{Encoding natural co-occurrences with controlled projection heads}\label{app:controlled}
Here, we verify whether our findings of \sectionautorefname~\ref{sec:cooccur} generalize to models trained with various projection heads. In \tableautorefname~\ref{tab:allresults}, we see that at least one internal layer in the projection head presents improved object co-occurrence encoding, compared to the final layer of the visual backbone. We also observe that this does not necessarily hold when the projection head is composed of a single linear layer (*-1 methods). This confirms our previous findings. We also see that, the deeper the projection head, the deeper the optimal layer for encoding object co-occurrences. This trend is consistent across all methods, datasets and metrics, with a single exception (BYOL-4 for COCO-THINGS RSA Kendall's tau). Overall, this shows that a representational structure capturing object co-occurrences tend to appear in deeper layers of the projection, though not the final one.

\begin{table}
\caption{Object Co-occurrences encoding metrics for 3 models (method-n) where n represents the number of linear layers in the projection head.}
\label{tab:allresults}
\begin{tabular}{c|ccccc|ccccc}
\toprule
\multirow{2}{*}{Method}& \multicolumn{5}{|c|}{RSA Kendall's tau} & \multicolumn{5}{|c}{Context accuracy} \\
 & B & mP.0 & mP.1 & mP.2 & mP.3  & B & P.0 & P.1 & P.2 & P.3  \\
\midrule
     \multicolumn{11}{c}{COCO-THINGS}\\
\midrule
 SimCLR-1 & \textbf{4.3} & -0.4 & na & na & na  & \textbf{36.6} & 35.3 & na & na & na  \\
 SimCLR-2 & 6.3 & \textbf{14.2} & -0.6 & na & na & 38.4 & \textbf{42.4} & 38.1 & na & na \\
 SimCLR-3 & 4.0 & \textbf{13.3} & 9.3 & -1.9 & na  & 36.5 & \textbf{40.7} & 39.6 & 34.8 & na \\
SimCLR-4 & 4.5 & \textbf{10.6} & 7.3 & 8.9 & 0.9  & 29.3 & 35.4 & 35.6 & \textbf{38.6} & 34.8  \\
\midrule
BYOL-1 & -1.9 & \textbf{2.3} & na & na & na  & 34.0 & \textbf{37.6} & na & na & na  \\
BYOL-2 & 2.2 & \textbf{22.5} & -3.4 & na & na  & 35.2 & \textbf{46.8} & 35.6 & na & na\\
BYOL-3 & 5.3 & 12.0 & \textbf{14.1} & -1.9 & na  & 36.4 & \textbf{42.7} & 41.8 & 33.7 & na \\
BYOL-4 & 6.1 & \textbf{15.9} & 15.7 & 13.7 & 4.3 & 39.7 & 42.8 & 40.1 & \textbf{43.8} & 35.1 \\
\midrule
DINO-1 & 7.0 & \textbf{12.8} & na & na & na & \textbf{36.9} & 30.4 & na & na & na \\
DINO-2 & -2.9 & \textbf{21.3} & 8.6 & na & na  & 24.9 & \textbf{41.0} & 35.3 &  na & na \\
DINO-3 & -3.4 & 18.4 & \textbf{21.4} & 7.8 & na  & 26.8 & \textbf{41.7} & 41.4 & 30.3 &  na \\
DINO-4 & -3.5 & 12.2 & 12.7 & \textbf{20.6} & 9.0  & 23.5 & 37.5 & \textbf{38.5} & 36.0 & 33.7 \\
\midrule
\multicolumn{11}{c}{ADE20k-THINGS}\\
\midrule
SimCLR-1 & \textbf{12.4} & 11.5 & na & na & na  & 23.7 & \textbf{24.5} & na & na & na  \\
SimCLR-2 & 11.7 & \textbf{19.7} & 12.7 & na & na & 20.6 & \textbf{22.6} & 22.2 & na & na \\
SimCLR-3 & 8.8 & 15.6 & \textbf{18.6} & 12.2 & na  & 20.3 & 23.3 & \textbf{24.4} & 22.8 & na \\
SimCLR-4 & 8.2 & 17.0 & \textbf{20.5} & 18.2 & 13.4  & 23.0 & 24.4 & \textbf{25.9} & 24.6 & 25.4 \\
\midrule
BYOL-1 & 10.6 & \textbf{13.8} & na & na & na  & 19.9 & \textbf{22.3} & na & na & na  \\
BYOL-2 & 9.3 & \textbf{16.4} & 11.5 & na & na  & 20.5 & \textbf{23.1} & 20.7 & na & na  \\
BYOL-3 & 9.0 & 17.7 & \textbf{24.4} & 10.2 & na  & 20.2 & 21.0 & \textbf{24.0} & 21.5 & na  \\
BYOL-4 & 8.7 & 18.1 & \textbf{23.4} & 19.1 & 16.9  & 19.9 & 20.6 & 21.3 & \textbf{25.3} & 22.4  \\
\midrule
DINO-1 & \textbf{16.1} & 14.0 & na & na & na  & 24.0 & \textbf{26.4} & na & na & na  \\
DINO-2 & 3.7 & \textbf{15.4} & 14.7 & na& na  & 12.6 & \textbf{27.3} & 25.6 & na & na \\
DINO-3 & 3.2 & 17.0 & \textbf{19.9} & 14.2 & na  & 13.1 & 24.9 & \textbf{26.7} & 19.6 & na \\
DINO-4 & 2.8 & 17.0 & \textbf{17.4} & 16.9 & 13.5 & 13.7 & 24.7 & 25.9 & 26.0 & \textbf{27.7}  \\
\midrule
    \multicolumn{11}{c}{LVIS-LVIS} \\
\midrule
SimCLR-1 & \textbf{26.9} & 26.6 & na & na & na  & \textbf{18.3} & 17.2  & na & na & na  \\
SimCLR-2 & 33.5 & \textbf{38.3} & 26.0 & na & na  & 18.8 & \textbf{20.9} & 18.1 & na & na  \\
SimCLR-3 & 32.8 & \textbf{37.0} & 36.9 & 26.1 & na  & 18.1 & 19.8 & \textbf{20.3} & 18.0 & na  \\
SimCLR-4 & 26.7 & 31.4 & \textbf{33.2} & \textbf{33.2} & 25.2  & 17.3 & 18.8 & 19.4 & \textbf{19.7} & 17.9 \\
\midrule
BYOL-1 & 23.9 & \textbf{26.3} & na & na & na & \textbf{17.5} & 16.9  & na & na & na  \\
BYOL-2 & 33.5 & \textbf{40.7} & 21.7 & na & na  & 18.3 & \textbf{20.6} & 14.2 & na & na  \\
BYOL-3 & 37.1 & 40.9 & \textbf{42.0} & 24.5 & na  & 17.9 & 20.5 & \textbf{20.7} & 16.5 & na  \\
BYOL-4 & 38.6 & 41.0 & \textbf{42.7 }& 35.9 & 26.2  & 18.7 & 20.1 &\textbf{ 20.5} & 20.2 & 18.2  \\
\midrule
DINO-1 & \textbf{39.3} & 27.4 &  na & na & na  & \textbf{18.8} & 17.7  & na & na & na  \\
DINO-2 & 41.1 & \textbf{43.4} & 25.7 & na & na  &19.1 & \textbf{20.6} & 16.8 & na & na  \\
DINO-3 & 40.1 & 42.4 & \textbf{42.9} & 26.8 & na  & 18.6 & 19.5 & \textbf{20.2} & 18.4 & na  \\
DINO-4 & 39.8 & 41.0 & 40.6 & \textbf{42.2} & 26.6  & 19.0 & 19.6 & 19.5 & \textbf{20.7} & 18.4   \\
\bottomrule
\end{tabular}
\end{table}

\begin{table}
\caption{Image-level object co-occurrences encoding metrics for 3 models (method-n) where n represents the number of linear layers in the projection head.}
\label{tab:imagesimilarity}
\begin{tabular}{c|ccccc|ccccc}
\toprule
\multirow{2}{*}{Method}& \multicolumn{5}{|c|}{RSA Kendall's tau} & \multicolumn{5}{|c}{Context accuracy} \\
 & B & mP.0 & mP.1 & mP.2 & mP.3  & B & P.0 & P.1 & P.2 & P.3  \\
\midrule
     \multicolumn{11}{c}{COCO-THINGS}\\
\midrule
SimCLR-1 & \textbf{0.5} & -4.1 & na & na & na & \textbf{27.6} & 26.4 & na & na & na \\
SimCLR-2 & 1.0 & \textbf{9.0} & -4.1 & na & na & 29.1 & \textbf{31.3} & 28.5 & na & na \\
SimCLR-3 & -1.2 & \textbf{6.6} & 4.5 & -5.1 & na & 27.6 & \textbf{30.3} & 30.1 & 25.7 & na  \\
SimCLR-4 & \textbf{9.5} & -1.2 & 6.6 & 4.5 & -5.1 & 25.1 & 27.8 & 27.6 & \textbf{29.6} & 27.1 \\
\midrule
BYOL-1 & -4.1 & \textbf{-0.3} & na & na & na & 25.5 & \textbf{26.4} & na & na & na  \\
BYOL-2 & -2.4 & \textbf{20.1} & -5.0 & na & na & 28.9 & \textbf{32.9} & 22.2 & na & na  \\
BYOL-3 & 1.2 & 6.7 & \textbf{11.5} & -3.7 & na & 29.4 & 31.5 & \textbf{32.7} & 23.3 & na \\
BYOL-4 & 2.3 & 13.2 & \textbf{15.9} & 13.6 & 4.0 & 32.4 & 32.7 & 33.3 & \textbf{34.0} & 29.4 \\
\midrule
DINO-1 & -2.0 & \textbf{16.1 }& na & na & na & 16.6 & \textbf{25.4}  & na & na & na \\
DINO-2 & -1.7 & \textbf{20.9} & 10.5 & na & na & 17.1 & \textbf{29.8} & 24.5  & na & na \\
DINO-3 & -2.7 & 18.6 & \textbf{23.4} & 9.7 & na & 17.2 & 30.9 & \textbf{33.3} & 26.6  & na \\
DINO-4 & -2.2 & 10.7 & 14.7 & \textbf{27.1} & 13.1 & 17.2 & 30.0 & 29.6 & \textbf{30.1} & 26.2  \\
\midrule
\multicolumn{11}{c}{ADE20k-THINGS}\\
\midrule
SimCLR-1 &  5.3 & \textbf{5.5} & na & na & na  & \textbf{20.3} & 19.1 & na & na & na  \\
SimCLR-2 & 2.4 & \textbf{12.8} & 7.0 & na & na  & 17.8 & \textbf{20.3} & 18.8 & na & na \\
SimCLR-3 & -0.9 & 7.3 & \textbf{12.5} & 6.4 & na & 17.0 & 17.9 & \textbf{20.6} & 18.6 & na \\
SimCLR-4 & 0.1 & 8.1 & \textbf{12.5} & 11.6 & 7.7 & 16.8 & 17.9 & 19.2 & \textbf{20.1} & 18.2 \\
\midrule
BYOL-1 & 5.5 & \textbf{9.9} & na & na & na & \textbf{16.6} & 16.1 & na & na & na  \\
BYOL-2 & 0.8 & 7.1 & \textbf{9.2} & na & na & 17.4 & \textbf{18.7} & 17.2 & na & na  \\
BYOL-3 & -1.1 & 10.3 & \textbf{19.9} & 8.1 & na & 18.0 & 16.9 & \textbf{21.0} & 17.7 & na  \\
BYOL-4 & -0.9 & 10.7 & \textbf{16.7} & 14.5 & 15.9 & 18.2 & 18.2 & \textbf{20.3} & 17.7 & 18.9  \\
\midrule
DINO-1 & 1.5 & \textbf{11.7} & na & na & na & 10.3 & \textbf{20.5} & na & na & na \\
DINO-2 & 1.6 & \textbf{10.9} & 11.5 & na & na & 10.6 & 19.6 & \textbf{19.9} & na & na \\
DINO-3 & 1.3 & 5.5 & \textbf{11.5} & 10.7 & na & 11.6 & 18.1 & \textbf{19.6} & 18.7  & na \\
DINO-4 & 1.0 & 8.1 & 7.6 & 8.2 & \textbf{10.4} & 10.8 & 18.6 & 19.1 & 20.9 & \textbf{21.5 } \\
\midrule
    \multicolumn{11}{c}{LVIS-LVIS} \\
\midrule
SimCLR-1 & 1.4 & \textbf{2.2} & na & na & na & \textbf{14.3} & 13.1 & na & na & na \\
SimCLR-2 & 3.8 & \textbf{11.7} & 1.7 & na & na & 14.1 & \textbf{15.2} & 13.6& na & na  \\
SimCLR-3 & -0.9 & 3.8 & \textbf{8.1 }& 1.8 & na & 13.7 & \textbf{14.9} & \textbf{14.9} & 13.5 & na \\
SimCLR-4 & 1.1 & 2.8 & 6.1 & \textbf{7.9} & 3.6  & 12.9 & 14.2 & 14.2 & \textbf{14.6} & 13.2  \\
\midrule
BYOL-1 & 0.7 & \textbf{2.3} & na & na & na & \textbf{14.0} & 13.8 & na & na & na  \\
BYOL-2 & 0.5 & \textbf{11.0} & 0.5 & na & na  & 14.0 & \textbf{15.7} & 12.3 & na & na \\
BYOL-3 & 4.8 & 7.6 & \textbf{11.1} & 0.2 & na  & 14.0 & \textbf{15.8} & 15.5 & 14.0 & na  \\
BYOL-4 & 4.7 & 9.1 & \textbf{11.8} & 10.8 & 5.3 & 15.1 & \textbf{16.1} & 15.4 & 15.2 & 15.2  \\
\midrule
DINO-1 & -0.0 & \textbf{14.1} & na & na & na & \textbf{15.1} & 14.4 & na & na & na \\
DINO-2 & -0.2 & \textbf{10.3 }& 14.6 & na & na & \textbf{15.5} & 15.3 & 14.8  & na & na \\
DINO-3 & -0.6 & 2.3 & 12.2 & \textbf{14.9} & na & 14.7 & 15.0 & \textbf{15.8} & 14.4  & na \\
DINO-4 & 0.8 & 5.4 & -1.2 & 14.4 & \textbf{15.0} & 14.7 & 15.2 & 15.4 & \textbf{16.6} & 14.9 \\
\bottomrule
\end{tabular}
\end{table}

\subsection{Co-occurrence encoding metrics based on image-level comparisons}\label{app:imagelevel}

In our method (\sectionautorefname~\ref{sec:datasets}), we built category representations to derive the representational dissimilarity matrices of models' layers. This tends to suppress intra-category variations between images. Here, we check whether our conclusions hold when we discard this and directly compute similarities between image pairs, which are thereafter averaged across pairs of categories. To derive a context accuracy, we perform one-shot context classification across 2,000, 2,000 and 10,000 tuples of images, instead of tuples of category prototypes, for COCO-THINGS, ADE20k-THINGS and LVIS-LVIS, respectively. In \tableautorefname~\ref{tab:imagesimilarity}, we see a trend consistent with our previous results: with a non-linear projection head, the projection head consistently outperforms the last layer of the backbone in encoding object co-occurrences. However, the results are generally lower than with our main evaluation protocol (\appendixautorefname~\ref{tab:allresults}), indicating layers encodes features that vary across images from the same categories, which impedes category co-occurrence encoding.

\subsection{Evaluation of more hidden layers}\label{app:morehidden}

\begin{figure}
    \centering
    \includegraphics[width=1\linewidth]{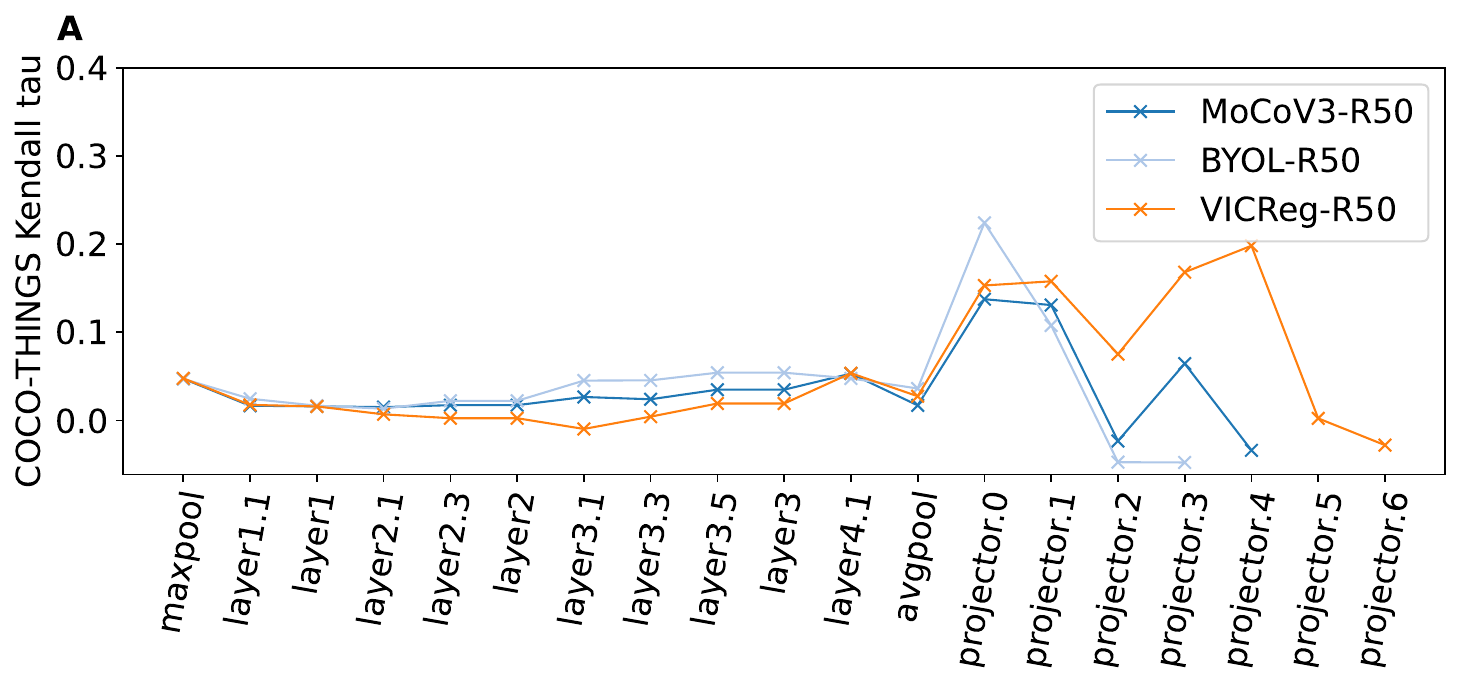}
    \includegraphics[width=1\linewidth]{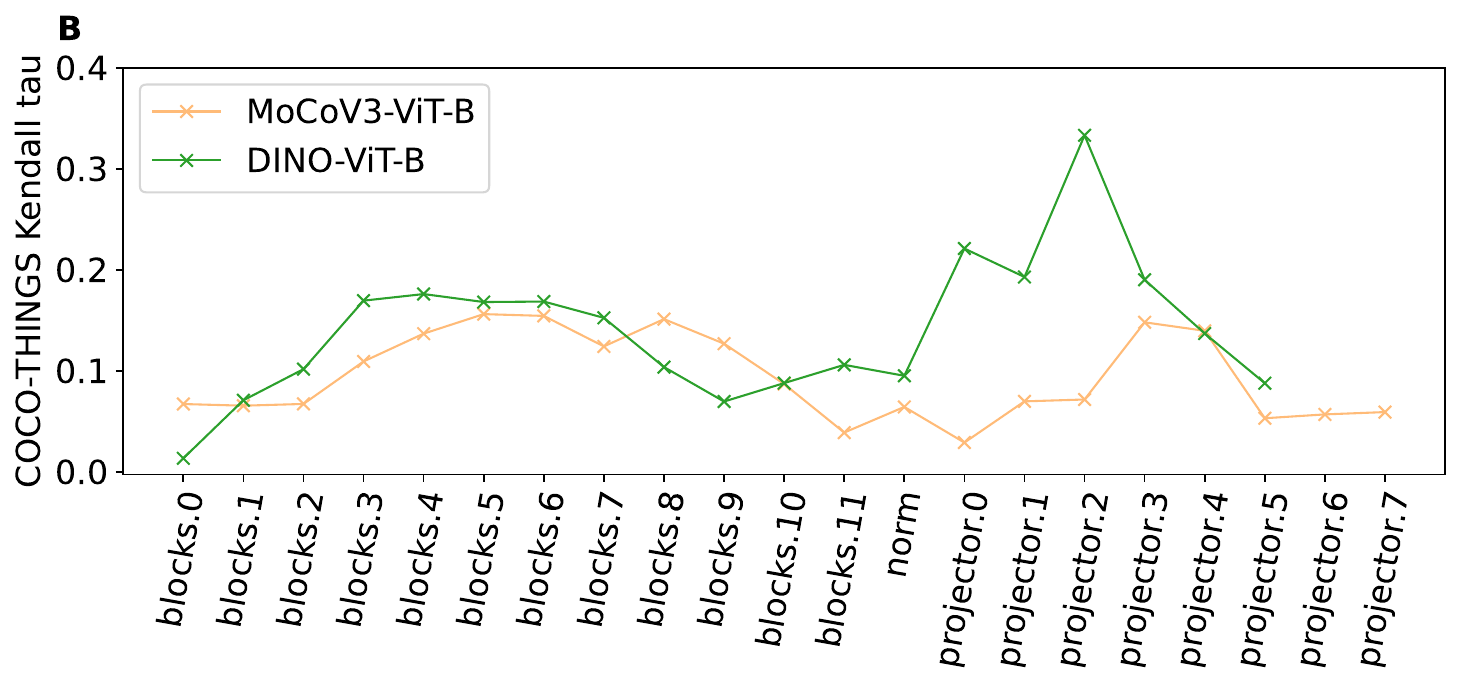}
    \includegraphics[width=1\linewidth]{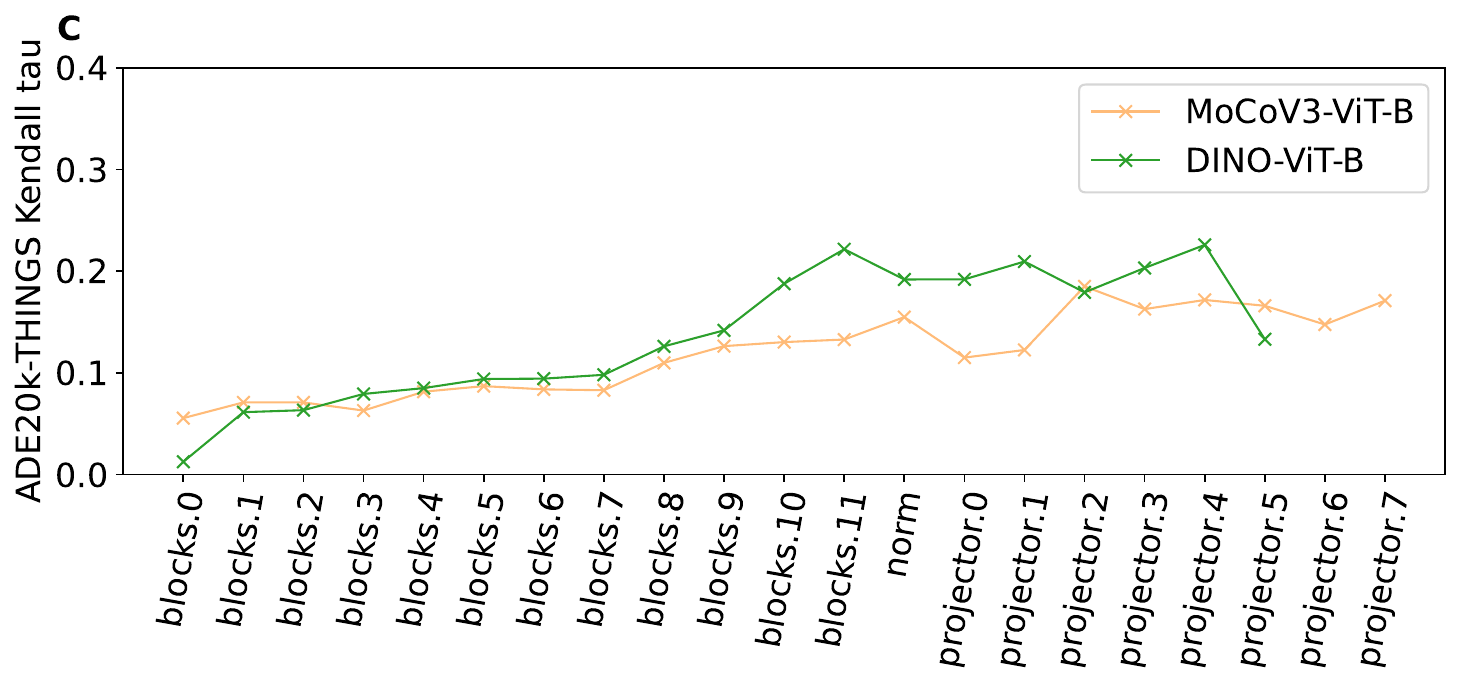}
    \caption{RSA Kendall's tau for 12 layers prior to the projection head, and layers in the projection head for the five main pre-trained models considered in our study. We consider A) COCO-THINGS with ResNet50 models, B) COCO-THINGS with ViT models and C) ADE20k-THINGS with ViT models.}
    \label{fig:alllayers}
\end{figure}

We found in \sectionautorefname~\ref{sec:cooccur} that the projection head learns representations that better model natural object co-occurrence, compared to the final layer of the visual backbone. Here, we extend our analysis to consider more layers within the neural network model. For internal layers, we take the spatial average of features for the ResNet50, and the register token for Vision Transformers. For the latter, we found it to work better than spatial average in general. In \figureautorefname~\ref{fig:alllayers}A-B), we see that Vision Transformers and ResNets progressively encode inter-object semantic relationships in a different manner. ResNets make little progress through the processing hierarchy for encoding object co-occurrences. In contrast, Vision Transformers exhibit a rapid increase in correlation, before dropping and increasing again with the projection head. However, this specific pattern does not old for ADE20k (\figureautorefname~\ref{fig:alllayers}C), suggesting that this may be due to the specific images of COCO-THINGS. Overall, these results confirm the superiority of layers in the projection head for encoding object co-occurrences.

\subsection{ADE20k-THINGS in-context category classification}\label{app:hierarchy2}

In \sectionautorefname~\ref{sec:hierarchy}, we showed that the in-context category classification decreases along deeper layers of the projection head, for THINGS-COCO and LVIS-LVIS setups. Here, we analyze results for the THINGS-ADE20k setup. In \tableautorefname~\ref{tab:incontext2}, we observe that the accuracy generally decreases across layers of the projection head (BYOL-R50, VICReg-R50), confirming our prior findings. However, the accuracy drop is less pronounced than THINGS-COCO and LVIS-LVIS (\sectionautorefname~\ref{sec:hierarchy}) and remains stable in some cases (MoCoV3-ViT-B).


\begin{table}
\centering
\caption{In-context one-shot category classification accuracy. We show average accuracy $\pm$ standard deviation, computed across three context assignments.}
\label{tab:incontext2}
\begin{tabular}{l|ccccc}
\toprule
Method & B & P.0 & P.1 & P.2 & P.3 \\
\midrule
MoCoV3-R50 & \textbf{91.9 $\pm$ 0.56} & 84.4 $\pm$ 0.87 & 83.7 $\pm$ 0.57 & na & na \\
BYOL-R50 & \textbf{90.2 $\pm$ 0.63} & 86.1 $\pm$ 0.32 & 73.5 $\pm$ 1.09 & na & na \\
VICReg-R50 & \textbf{89.4 $\pm$ 0.96} & 85.8 $\pm$ 0.69 & 83.8 $\pm$ 0.83 & 80.0 $\pm$ 1.35 & na \\
MoCoV3-ViT-B & 91.1 $\pm$ 1.00 & \textbf{91.3 $\pm$ 1.25} & 88.3 $\pm$ 0.65 & 90.3 $\pm$ 1.12 & na \\
DINO-ViT-B & \textbf{94.6 $\pm$ 0.18} & 93.6 $\pm$ 0.23 & 90.7 $\pm$ 0.45 & 94.3 $\pm$ 0.16 & 89.7 $\pm$ 0.71 \\
\bottomrule
\end{tabular}
\end{table}

\subsection{Evaluation of foundational backbones}\label{app:foundational}

Despite the absence of publicly released weights in foundational models, we report the performance of CLIP \citep{radford2021learning}, DinoV2 \citep{oquab2024dinov2} and a supervised resnet50 trained on Imagenet-1k \citep{he2016deep} in encoding object co-occurrences. Interestingly, we find that the projection head of DINO rivals with models that benefit supervision or bigger image datasets. The projection head is crucial to bridge the gap between the backbone of DINO and backbones of other methods.

\begin{table}
\centering
\label{tab:foundational}
\caption{Object co-occurrence encoding metrics for each model.}
\begin{tabular}{c|c|c}
\toprule
Method & RSA Kendall's tau & Context accuracy \\
\midrule
     \multicolumn{3}{c}{COCO-THINGS}\\
\midrule
DINO-ViT-B (backbone) & 9.0 & 43.5 \\
DINO-ViT-B (projection head) & \textbf{33.4} & 50.3 \\
CLIP-ViT-L (backbone) &  28.1 & \textbf{58.2} \\
DINOv2-ViT-L (backbone) & 22.4 & 37.6 \\
Supervised R50 (backbone) & 10.6 & 36.6 \\
\midrule
     \multicolumn{3}{c}{ADE20k-THINGS}\\
\midrule
DINO-ViT-B (backbone)&19.3 & 22.4 \\
DINO-ViT-B (projection head) & 22.4 & \textbf{33.0} \\
CLIP-ViT-L (backbone) & \textbf{24.7} & 28.1 \\
DINOv2-ViT-L (backbone) &20.4 & 29.9 \\
Supervised R50 (backbone) &23.5 & 22.6 \\
\midrule
     \multicolumn{3}{c}{LVIS-LVIS}\\
\midrule
DINO-ViT-B (backbone) & 39.8 & 21.4 \\
DINO-ViT-B (projection head) &  43.3 & 23.9 \\
CLIP-ViT-L (backbone)  & \textbf{44.4} & 21.9 \\
DINOv2-ViT-L (backbone) & 41.5 & \textbf{24.3} \\
Supervised R50 (backbone) & 37.0 & 20.3 \\
\bottomrule
\end{tabular}
\end{table}
\newpage
\section*{NeurIPS Paper Checklist}

\begin{enumerate}

\item {\bf Claims}
    \item[] Question: Do the main claims made in the abstract and introduction accurately reflect the paper's contributions and scope?
    \item[] Answer: \answerYes{} 
    \item[] Justification: All claims are supported by our experiments.
    \item[] Guidelines:
    \begin{itemize}
        \item The answer \answerNA{} means that the abstract and introduction do not include the claims made in the paper.
        \item The abstract and/or introduction should clearly state the claims made, including the contributions made in the paper and important assumptions and limitations. A \answerNo{} or \answerNA{} answer to this question will not be perceived well by the reviewers. 
        \item The claims made should match theoretical and experimental results, and reflect how much the results can be expected to generalize to other settings. 
        \item It is fine to include aspirational goals as motivation as long as it is clear that these goals are not attained by the paper. 
    \end{itemize}

\item {\bf Limitations}
    \item[] Question: Does the paper discuss the limitations of the work performed by the authors?
    \item[] Answer: \answerYes{} 
    \item[] Justification: We have included a paragraph about limitations in the Conclusion section.
    \item[] Guidelines:
    \begin{itemize}
        \item The answer \answerNA{} means that the paper has no limitation while the answer \answerNo{} means that the paper has limitations, but those are not discussed in the paper. 
        \item The authors are encouraged to create a separate ``Limitations'' section in their paper.
        \item The paper should point out any strong assumptions and how robust the results are to violations of these assumptions (e.g., independence assumptions, noiseless settings, model well-specification, asymptotic approximations only holding locally). The authors should reflect on how these assumptions might be violated in practice and what the implications would be.
        \item The authors should reflect on the scope of the claims made, e.g., if the approach was only tested on a few datasets or with a few runs. In general, empirical results often depend on implicit assumptions, which should be articulated.
        \item The authors should reflect on the factors that influence the performance of the approach. For example, a facial recognition algorithm may perform poorly when image resolution is low or images are taken in low lighting. Or a speech-to-text system might not be used reliably to provide closed captions for online lectures because it fails to handle technical jargon.
        \item The authors should discuss the computational efficiency of the proposed algorithms and how they scale with dataset size.
        \item If applicable, the authors should discuss possible limitations of their approach to address problems of privacy and fairness.
        \item While the authors might fear that complete honesty about limitations might be used by reviewers as grounds for rejection, a worse outcome might be that reviewers discover limitations that aren't acknowledged in the paper. The authors should use their best judgment and recognize that individual actions in favor of transparency play an important role in developing norms that preserve the integrity of the community. Reviewers will be specifically instructed to not penalize honesty concerning limitations.
    \end{itemize}

\item {\bf Theory assumptions and proofs}
    \item[] Question: For each theoretical result, does the paper provide the full set of assumptions and a complete (and correct) proof?
    \item[] Answer: \answerNA{} 
    \item[] Justification:
    \item[] Guidelines:
    \begin{itemize}
        \item The answer \answerNA{} means that the paper does not include theoretical results. 
        \item All the theorems, formulas, and proofs in the paper should be numbered and cross-referenced.
        \item All assumptions should be clearly stated or referenced in the statement of any theorems.
        \item The proofs can either appear in the main paper or the supplemental material, but if they appear in the supplemental material, the authors are encouraged to provide a short proof sketch to provide intuition. 
        \item Inversely, any informal proof provided in the core of the paper should be complemented by formal proofs provided in appendix or supplemental material.
        \item Theorems and Lemmas that the proof relies upon should be properly referenced. 
    \end{itemize}

    \item {\bf Experimental result reproducibility}
    \item[] Question: Does the paper fully disclose all the information needed to reproduce the main experimental results of the paper to the extent that it affects the main claims and/or conclusions of the paper (regardless of whether the code and data are provided or not)?
    \item[] Answer: \answerYes{} 
    \item[] Justification: The method section accurately describes or analysis protocol. \appendixautorefname~\ref{app:newlypretrained} describes the hyper-parameters used to pretrain our models.
    \item[] Guidelines:
    \begin{itemize}
        \item The answer \answerNA{} means that the paper does not include experiments.
        \item If the paper includes experiments, a \answerNo{} answer to this question will not be perceived well by the reviewers: Making the paper reproducible is important, regardless of whether the code and data are provided or not.
        \item If the contribution is a dataset and\slash or model, the authors should describe the steps taken to make their results reproducible or verifiable. 
        \item Depending on the contribution, reproducibility can be accomplished in various ways. For example, if the contribution is a novel architecture, describing the architecture fully might suffice, or if the contribution is a specific model and empirical evaluation, it may be necessary to either make it possible for others to replicate the model with the same dataset, or provide access to the model. In general. releasing code and data is often one good way to accomplish this, but reproducibility can also be provided via detailed instructions for how to replicate the results, access to a hosted model (e.g., in the case of a large language model), releasing of a model checkpoint, or other means that are appropriate to the research performed.
        \item While NeurIPS does not require releasing code, the conference does require all submissions to provide some reasonable avenue for reproducibility, which may depend on the nature of the contribution. For example
        \begin{enumerate}
            \item If the contribution is primarily a new algorithm, the paper should make it clear how to reproduce that algorithm.
            \item If the contribution is primarily a new model architecture, the paper should describe the architecture clearly and fully.
            \item If the contribution is a new model (e.g., a large language model), then there should either be a way to access this model for reproducing the results or a way to reproduce the model (e.g., with an open-source dataset or instructions for how to construct the dataset).
            \item We recognize that reproducibility may be tricky in some cases, in which case authors are welcome to describe the particular way they provide for reproducibility. In the case of closed-source models, it may be that access to the model is limited in some way (e.g., to registered users), but it should be possible for other researchers to have some path to reproducing or verifying the results.
        \end{enumerate}
    \end{itemize}

\item {\bf Open access to data and code}
    \item[] Question: Does the paper provide open access to the data and code, with sufficient instructions to faithfully reproduce the main experimental results, as described in supplemental material?
    \item[] Answer: \answerNA{} 
    \item[] Justification: We plan to release our documented code upon acceptance.
    \item[] Guidelines:
    \begin{itemize}
        \item The answer \answerNA{} means that paper does not include experiments requiring code.
        \item Please see the NeurIPS code and data submission guidelines (\url{https://neurips.cc/public/guides/CodeSubmissionPolicy}) for more details.
        \item While we encourage the release of code and data, we understand that this might not be possible, so \answerNo{} is an acceptable answer. Papers cannot be rejected simply for not including code, unless this is central to the contribution (e.g., for a new open-source benchmark).
        \item The instructions should contain the exact command and environment needed to run to reproduce the results. See the NeurIPS code and data submission guidelines (\url{https://neurips.cc/public/guides/CodeSubmissionPolicy}) for more details.
        \item The authors should provide instructions on data access and preparation, including how to access the raw data, preprocessed data, intermediate data, and generated data, etc.
        \item The authors should provide scripts to reproduce all experimental results for the new proposed method and baselines. If only a subset of experiments are reproducible, they should state which ones are omitted from the script and why.
        \item At submission time, to preserve anonymity, the authors should release anonymized versions (if applicable).
        \item Providing as much information as possible in supplemental material (appended to the paper) is recommended, but including URLs to data and code is permitted.
    \end{itemize}

\item {\bf Experimental setting/details}
    \item[] Question: Does the paper specify all the training and test details (e.g., data splits, hyperparameters, how they were chosen, type of optimizer) necessary to understand the results?
    \item[] Answer: \answerYes{} 
    \item[] Justification: \appendixautorefname~\ref{app:newlypretrained} describes the hyper-parameters used to pre-train our models.
    \item[] Guidelines:
    \begin{itemize}
        \item The answer \answerNA{} means that the paper does not include experiments.
        \item The experimental setting should be presented in the core of the paper to a level of detail that is necessary to appreciate the results and make sense of them.
        \item The full details can be provided either with the code, in appendix, or as supplemental material.
    \end{itemize}

\item {\bf Experiment statistical significance}
    \item[] Question: Does the paper report error bars suitably and correctly defined or other appropriate information about the statistical significance of the experiments?
    \item[] Answer: \answerYes{} 
    \item[] Justification: \appendixautorefname~\ref{app:std} reports the missing standard deviation in our experiments. For clarity and compactness, we do not report it for all our results.
    \item[] Guidelines:
    \begin{itemize}
        \item The answer \answerNA{} means that the paper does not include experiments.
        \item The authors should answer \answerYes{} if the results are accompanied by error bars, confidence intervals, or statistical significance tests, at least for the experiments that support the main claims of the paper.
        \item The factors of variability that the error bars are capturing should be clearly stated (for example, train/test split, initialization, random drawing of some parameter, or overall run with given experimental conditions).
        \item The method for calculating the error bars should be explained (closed form formula, call to a library function, bootstrap, etc.)
        \item The assumptions made should be given (e.g., Normally distributed errors).
        \item It should be clear whether the error bar is the standard deviation or the standard error of the mean.
        \item It is OK to report 1-sigma error bars, but one should state it. The authors should preferably report a 2-sigma error bar than state that they have a 96\% CI, if the hypothesis of Normality of errors is not verified.
        \item For asymmetric distributions, the authors should be careful not to show in tables or figures symmetric error bars that would yield results that are out of range (e.g., negative error rates).
        \item If error bars are reported in tables or plots, the authors should explain in the text how they were calculated and reference the corresponding figures or tables in the text.
    \end{itemize}

\item {\bf Experiments compute resources}
    \item[] Question: For each experiment, does the paper provide sufficient information on the computer resources (type of compute workers, memory, time of execution) needed to reproduce the experiments?
    \item[] Answer: \answerYes{} 
    \item[] Justification: We give information about our hardware and execution time in \appendixautorefname~\ref{app:newlypretrained}
    \item[] Guidelines:
    \begin{itemize}
        \item The answer \answerNA{} means that the paper does not include experiments.
        \item The paper should indicate the type of compute workers CPU or GPU, internal cluster, or cloud provider, including relevant memory and storage.
        \item The paper should provide the amount of compute required for each of the individual experimental runs as well as estimate the total compute. 
        \item The paper should disclose whether the full research project required more compute than the experiments reported in the paper (e.g., preliminary or failed experiments that didn't make it into the paper). 
    \end{itemize}
    
\item {\bf Code of ethics}
    \item[] Question: Does the research conducted in the paper conform, in every respect, with the NeurIPS Code of Ethics \url{https://neurips.cc/public/EthicsGuidelines}?
    \item[] Answer: \answerYes{} 
    \item[] Justification: We have verified each point of the code of Ethics.
    \item[] Guidelines:
    \begin{itemize}
        \item The answer \answerNA{} means that the authors have not reviewed the NeurIPS Code of Ethics.
        \item If the authors answer \answerNo, they should explain the special circumstances that require a deviation from the Code of Ethics.
        \item The authors should make sure to preserve anonymity (e.g., if there is a special consideration due to laws or regulations in their jurisdiction).
    \end{itemize}

\item {\bf Broader impacts}
    \item[] Question: Does the paper discuss both potential positive societal impacts and negative societal impacts of the work performed?
    \item[] Answer: \answerNA{} 
    \item[] Justification: This work is an analysis of existing methods. There is no direct societal impact.
    \item[] Guidelines:
    \begin{itemize}
        \item The answer \answerNA{} means that there is no societal impact of the work performed.
        \item If the authors answer \answerNA{} or \answerNo, they should explain why their work has no societal impact or why the paper does not address societal impact.
        \item Examples of negative societal impacts include potential malicious or unintended uses (e.g., disinformation, generating fake profiles, surveillance), fairness considerations (e.g., deployment of technologies that could make decisions that unfairly impact specific groups), privacy considerations, and security considerations.
        \item The conference expects that many papers will be foundational research and not tied to particular applications, let alone deployments. However, if there is a direct path to any negative applications, the authors should point it out. For example, it is legitimate to point out that an improvement in the quality of generative models could be used to generate Deepfakes for disinformation. On the other hand, it is not needed to point out that a generic algorithm for optimizing neural networks could enable people to train models that generate Deepfakes faster.
        \item The authors should consider possible harms that could arise when the technology is being used as intended and functioning correctly, harms that could arise when the technology is being used as intended but gives incorrect results, and harms following from (intentional or unintentional) misuse of the technology.
        \item If there are negative societal impacts, the authors could also discuss possible mitigation strategies (e.g., gated release of models, providing defenses in addition to attacks, mechanisms for monitoring misuse, mechanisms to monitor how a system learns from feedback over time, improving the efficiency and accessibility of ML).
    \end{itemize}
    
\item {\bf Safeguards}
    \item[] Question: Does the paper describe safeguards that have been put in place for responsible release of data or models that have a high risk for misuse (e.g., pre-trained language models, image generators, or scraped datasets)?
    \item[] Answer: \answerNA{} 
    \item[] Justification:
    \item[] Guidelines:
    \begin{itemize}
        \item The answer \answerNA{} means that the paper poses no such risks.
        \item Released models that have a high risk for misuse or dual-use should be released with necessary safeguards to allow for controlled use of the model, for example by requiring that users adhere to usage guidelines or restrictions to access the model or implementing safety filters. 
        \item Datasets that have been scraped from the Internet could pose safety risks. The authors should describe how they avoided releasing unsafe images.
        \item We recognize that providing effective safeguards is challenging, and many papers do not require this, but we encourage authors to take this into account and make a best faith effort.
    \end{itemize}

\item {\bf Licenses for existing assets}
    \item[] Question: Are the creators or original owners of assets (e.g., code, data, models), used in the paper, properly credited and are the license and terms of use explicitly mentioned and properly respected?
    \item[] Answer: \answerYes{} 
    \item[] Justification: We cite each dataset and code that we use.
    \item[] Guidelines:
    \begin{itemize}
        \item The answer \answerNA{} means that the paper does not use existing assets.
        \item The authors should cite the original paper that produced the code package or dataset.
        \item The authors should state which version of the asset is used and, if possible, include a URL.
        \item The name of the license (e.g., CC-BY 4.0) should be included for each asset.
        \item For scraped data from a particular source (e.g., website), the copyright and terms of service of that source should be provided.
        \item If assets are released, the license, copyright information, and terms of use in the package should be provided. For popular datasets, \url{paperswithcode.com/datasets} has curated licenses for some datasets. Their licensing guide can help determine the license of a dataset.
        \item For existing datasets that are re-packaged, both the original license and the license of the derived asset (if it has changed) should be provided.
        \item If this information is not available online, the authors are encouraged to reach out to the asset's creators.
    \end{itemize}

\item {\bf New assets}
    \item[] Question: Are new assets introduced in the paper well documented and is the documentation provided alongside the assets?
    \item[] Answer: \answerNA{} 
    \item[] Justification: 
    \item[] Guidelines:
    \begin{itemize}
        \item The answer \answerNA{} means that the paper does not release new assets.
        \item Researchers should communicate the details of the dataset\slash code\slash model as part of their submissions via structured templates. This includes details about training, license, limitations, etc. 
        \item The paper should discuss whether and how consent was obtained from people whose asset is used.
        \item At submission time, remember to anonymize your assets (if applicable). You can either create an anonymized URL or include an anonymized zip file.
    \end{itemize}

\item {\bf Crowdsourcing and research with human subjects}
    \item[] Question: For crowdsourcing experiments and research with human subjects, does the paper include the full text of instructions given to participants and screenshots, if applicable, as well as details about compensation (if any)? 
    \item[] Answer: \answerNA{} 
    \item[] Justification: 
    \item[] Guidelines:
    \begin{itemize}
        \item The answer \answerNA{} means that the paper does not involve crowdsourcing nor research with human subjects.
        \item Including this information in the supplemental material is fine, but if the main contribution of the paper involves human subjects, then as much detail as possible should be included in the main paper. 
        \item According to the NeurIPS Code of Ethics, workers involved in data collection, curation, or other labor should be paid at least the minimum wage in the country of the data collector. 
    \end{itemize}

\item {\bf Institutional review board (IRB) approvals or equivalent for research with human subjects}
    \item[] Question: Does the paper describe potential risks incurred by study participants, whether such risks were disclosed to the subjects, and whether Institutional Review Board (IRB) approvals (or an equivalent approval/review based on the requirements of your country or institution) were obtained?
    \item[] Answer: \answerNA{} 
    \item[] Justification: 
    \item[] Guidelines:
    \begin{itemize}
        \item The answer \answerNA{} means that the paper does not involve crowdsourcing nor research with human subjects.
        \item Depending on the country in which research is conducted, IRB approval (or equivalent) may be required for any human subjects research. If you obtained IRB approval, you should clearly state this in the paper. 
        \item We recognize that the procedures for this may vary significantly between institutions and locations, and we expect authors to adhere to the NeurIPS Code of Ethics and the guidelines for their institution. 
        \item For initial submissions, do not include any information that would break anonymity (if applicable), such as the institution conducting the review.
    \end{itemize}

\item {\bf Declaration of LLM usage}
    \item[] Question: Does the paper describe the usage of LLMs if it is an important, original, or non-standard component of the core methods in this research? Note that if the LLM is used only for writing, editing, or formatting purposes and does \emph{not} impact the core methodology, scientific rigor, or originality of the research, declaration is not required.
    \item[] Answer: \answerNA{} 
    \item[] Justification: 
    \item[] Guidelines:
    \begin{itemize}
        \item The answer \answerNA{} means that the core method development in this research does not involve LLMs as any important, original, or non-standard components.
        \item Please refer to our LLM policy in the NeurIPS handbook for what should or should not be described.
    \end{itemize}

\end{enumerate}

\end{document}